\documentclass[]{article}
\usepackage{proceed2e}
\usepackage{times}

\usepackage{times}
\usepackage{amsmath,amsfonts,amssymb,mathtools,mathrsfs,dsfont}
\usepackage{amsthm}
\usepackage{bm}
\usepackage{dblfloatfix}
\usepackage{graphicx}
\usepackage{subcaption}
\usepackage{ragged2e}
\usepackage{natbib}
\usepackage{hyperref}
\usepackage[normalem]{ulem}
\usepackage{enumitem}
\usepackage{array}
\usepackage{tabularx,colortbl}
\usepackage{multicol}
\usepackage{multirow}
\usepackage{hhline}
\usepackage{makecell}
\usepackage{tikz}

\definecolor{pale_blue}{rgb}{0.84,0.92,1}

\title{Deep Graphs}

\author{
{\bf Emmanouil Antonios Platanios} \\
Machine Learning Department \\
Carnegie Mellon University \\
Pittsburgh, PA \\
\small\ttfamily{e.a.platanios@cs.cmu.edu}
\And
{\bf Alex Smola} \\
AWS Machine Learning \\
Amazon \\
Palo Alto, CA \\
\small\ttfamily{smola@amazon.com}
}

\begin{document}

\maketitle

\begin{abstract}
  We propose an algorithm for deep learning on networks and graphs. It
  relies on the notion that many graph algorithms, such as
  PageRank, Weisfeiler-Lehman, or Message Passing can be expressed as
  iterative vertex updates. Unlike previous methods which rely on the
  ingenuity of the designer, Deep Graphs are adaptive to the
  estimation problem. Training and deployment are both efficient, since the
  cost is $O(|E| + |V|)$, where $E$ and $V$ are the sets of edges and
  vertices respectively. In short, we \emph{learn} the recurrent
  update functions rather than positing their specific functional
  form.
  This yields an algorithm that achieves excellent accuracy on both
  graph labeling and regression tasks. 
\end{abstract}

\section{INTRODUCTION}

Tasks which involve graph structures are abundant in machine learning
and data mining. As an example, social networks can be represented as
graphs with users being vertices and friendship relationships being
edges. In this case, we might want to classify users or make
predictions about missing information from their profiles. Many such
tasks involve learning a representation for the vertices and the edges
of the corresponding graph structures. This often turns out to be
difficult, requiring a lot of task-specific feature engineering. In this paper,
we propose a generic algorithm for learning such representations
jointly with the tasks. This algorithm requires no task-specific
engineering and can be used for a wide range of problems.

The key idea is to realize that many graph algorithms are defined in
terms of vertex updates. That is, vertex features are updated
iteratively, based on their own features and those of their
neighbors. For instance, the PageRank algorithm \citep{PagBriMotWin98}
updates vertex attributes based on the popularity of its parents. In
fact, there even exist numerous graph processing frameworks based on
the very idea that many graph algorithms are vertex centric
\citep{LowGonKyrBicetal10,Malewicz:2010:PSL:1807167.1807184}. 
Unlike prior work, we do not posit the
form of the vertex update function but instead, we learn it from data.

In particular, we define a {\em recurrent neural network (RNN)} over
the graph structure where the features of each vertex (or edge) are
computed as a function of the neighboring vertices' and edges'
features. We call the resulting model a {\em Deep Graph (DG)}. 
The vertex features can be used to perform multiple tasks
simultaneously, e.g.\ in a {\em multi-task learning} setting, and the
proposed algorithm is able to learn the functions that generate these
features and the functions that perform these tasks,
jointly. Furthermore, it is able to learn representations for graph
vertices and edges that can, subsequently, be used by other algorithms
to perform new tasks, such as in {\em transfer learning}. Finally,
apart from the graph structure, this algorithm is also able to use
attributes of the vertices and edges of the graph, rendering it even
more powerful in cases where such attributes are available.

\begin{figure}[th!]
	\centering
        \includegraphics[width=\columnwidth]{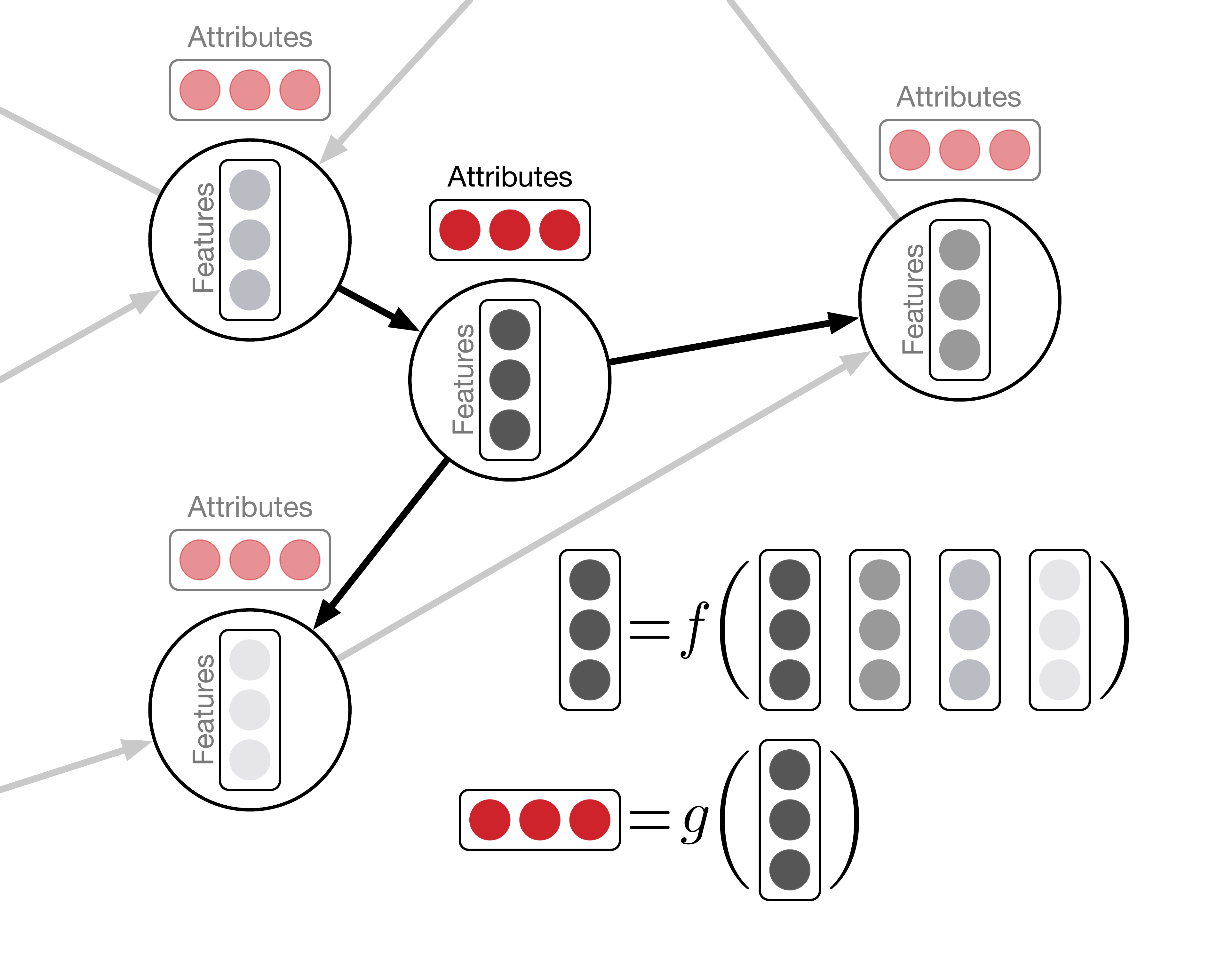}
	\caption{Vertex updates on a graph. Circles represent vertices,
          rectangles with colored circles inside them represent
          vectors, arrows represent edges, gray colored vectors
          correspond to the feature vectors that our method learns to
          compute as a function of the neighboring vertex feature
          vectors, and red and blue colored vectors correspond to
          vertex attributes that are either observed, or computed as a
          learned function of the corresponding feature vector.
	\label{fig:graph_figure}}
\end{figure}

Our work is arguably the first to take the notion of deep
\emph{networks} literally and to apply it to graphs and networks. To
some extent our work can be interpreted as that of learning the
general dynamics of brain ``neurons'' in the form of a mapping that takes
the vertex state and the neighbors' states and transforms their combination into
a new vertex state. By using sufficiently powerful latent state
dynamics we are able to capture nontrivial dependencies and address
the vanishing gradient problem that usually plagues recurrent
networks.

We performed several experiments showing the usefulness and power of DGs. Initially we attempted to learn PageRank and HITS (an algorithm also used for ranking vertices based on their popularity \citep{Kleinberg99}) in order to confirm that our approach can learn simple graph features. Our results indicate that DGs can learn to compute PageRank and HITS scores by only applying a vertex update equation only $6$ times, as opposed to hundreds or thousands of iterations that the original algorithms require. Furthermore, we also performed vertex classification experiments on $5$ diverse data sets, and DGs were shown to significantly outperform the current state-of-the-art.

\section{BACKGROUND}
\label{sec:background}

\paragraph{Notation.} We denote by $G=(V,E)$ a graph with vertices $V$
and edges $E$, i.e.\ for some vertices $i, j \in V$ we have
$(i,j) \in E$. Moreover, when available, we denote by $\psi(i)$ and
$\psi(i,j)$ attributes associated with vertices and edges,
respectively. We allow for directed graphs where $(i,j) \in E$ does
not imply $(j,i) \in E$, and we also allow for cases where
$\psi(i,j) \neq \psi(j,i)$. Furthermore, we denote the set of incoming
neighbors of vertex $i$ by $\mathcal{N}_{\textrm{in}}(i)$. That is,
the set of vertices such that for each vertex $j$ in that set, there
exists an edge $(j,i) \in E$. We denote the set of outgoing neighbors
of vertex $i$ by $\mathcal{N}_{\textrm{out}}(i)$. That is, the set of
vertices such that for each vertex $j$ in that set, there exists an
edge $(i,j) \in E$. Also, we denote the set of all neighbors of vertex
$i$ by
$\mathcal{N}(i) \triangleq \mathcal{N}_{\textrm{in}}(i) \cup
\mathcal{N}_{\textrm{out}}(i)$.

It is our goal to compute local vertex (and occasionally edge)
features $\phi(i)$ and $\phi(i,j)$ based on the graph structure $G$,
that can be used to estimate vertex and edge attributes $\psi(i)$ and
$\psi(i,j)$. In the next few paragraphs we review some existing
algorithms that do exactly that. Subsequently we show how our proposed
approach can be viewed as a more powerful generalization of those
algorithms that allows for more flexibility.

\paragraph{PageRank.} The most famous algorithm for attaching features
to a directed graph is arguably the {\em PageRank} algorithm of
\citet{PagBriMotWin98}. It aims to furnish vertices with a score
commensurate with their level of relevance. That is, the PageRank is
high for important pages and low for less relevant ones. The key idea
is that relevant pages relate to other relevant pages. Hence, a random
surfer traversing the web would be more likely to visit relevant
pages. A simple definition of the algorithm is to iterate the
following updates for $i\in\{1,\hdots,|V|\}$ until convergence:
\begin{equation}
\label{eq:pagerank_iteration}
	\pi_i \leftarrow \lambda \sum_{(i,j) \in E} |\mathcal{N}_\textrm{out}(j)|^{-1} \pi_j + (1-\lambda) |V|^{-1} \pi_i,
\end{equation}
Here $\pi_i$ is the PageRank score of vertex $i$, and $\lambda\in[0,1]$ is a damping factor. Note that in this case $\phi(i)=\pi_i$. We know that this iteration is contractive and thus it will converge quickly. To summarize, PageRank consists of repeatedly using vertex features $\phi(i)$ to recompute new vertex features based on a rather simple, yet ingenious iteration.

\paragraph{HITS:} The {\em HITS} algorithm of \citet{Kleinberg99} follows a very similar template. The key difference is that it computes two scores: {\em authority} and {\em hub}. Authority scores are computed using the hub scores of all  incoming neighbors of vertex $i$. Likewise, hub scores are computed using the authority scores of all outgoing neighbors of vertex $i$. This amounts to the following two iterations:
\begin{equation}
\label{eq:hits_iteration}
	\pi_{i} \leftarrow \sum_{(i,j) \in E} \rho_j,\quad\textrm{and}\quad\rho_{j} \leftarrow \sum_{(i,j) \in E} \pi_i,
\end{equation}
where $\pi_i$ is the hub score of vertex $i$ and $\rho_i$ is its authority score. It is easy to see that this iteration diverges and in order to avoid that, we normalize all authority and hub scores by the sum of squares of all authority and hub scores respectively, after each iteration. Note that in this case $\phi(i)=[\pi_i,\rho_i]$. Thus, HITS also consists of repeatedly using vertex features $\phi(i)$ to recompute new vertex features based on a simple iteration.

\paragraph{Weisfeiler-Lehman:} Another algorithm commonly used on unattributed graphs is the famous algorithm of \citet{WeiLeh68} which can generate sufficient conditions for graph isomorphism tests. Unlike PageRank and HITS, it is a discrete mapping that generates vertex features which can be used to uniquely identify vertices for suitable graphs. Its motivation is that if such an identification can be found, graph isomorphism becomes trivial since we now only need to check whether the sets of vertex features are identical. For simplicity of exposition, we assume, without loss of generality, that there exist collision-free hash functions\footnote{$2^{V}$ here denotes the power set of all vertices.} $h: 2^{V} \mapsto \mathbb{N}$. That is, we ignore the case of $h(i) = h(j)$ for $i \neq j$. Then, the algorithm consists of initializing $\phi(i) = 1$, for all $i \in V$, and subsequently performing iterations of the following equation, for $i\in\{1,\hdots,|V|\}$, until convergence:
\begin{equation}
\label{eq:weisfeler_lehman}
	\phi(i) \leftarrow h\left(\phi(i), \left\{\phi(j) \mid j \in \mathcal{N}(i)\right\}\right)
\end{equation}
In other words, the algorithm computes a new vertex hash based on the current hash and the hash of all of its neighbors. The algorithm terminates when the number of unique vertex features no longer increases. Note that whenever all vertex features $\phi(i)$ are unique, this immediately allows us to solve the graph isomorphism problem, since the iteration does not explicitly exploit the vertex index $i$. Also note that (\ref{eq:weisfeler_lehman}) can be extended trivially to graphs with vertex and edge attributes. This is simply accomplished by initializing $\phi(i) = \psi(i)$ for all $i \in V$ and setting\footnote{For brevity, we have slightly abused the notation for edge directionality here, but the concept described should remain clear.}:
\begin{equation}
	\phi(i) \leftarrow h\left(\phi(i), \left\{\phi(j), \psi(i,j) \mid j \in \mathcal{N}(i)\right\}\right)
\end{equation}
In other words, we use both vertex and edge attributes in generating
unique fingerprints (i.e., features) for the vertices. The
Weisfeiler-Lehman iteration is of interest in the current context,
because it can be used to generate useful features for graph vertices,
in general. In their prize-winning paper, \citet{SheBor10} use
iterations of this algorithm to obtain a set of vertex features,
ranging from the generic (at initialization time) to the highly
specific (at convergence time). They allow one to compare vertices and
perform estimation efficiently, since vertex hashes at iteration $k$
will be identical, whenever the $k$-step neighborhood of two vertices
is identical. Their algorithm is very fast and yields high quality
estimates on unattributed graphs. However, it has resisted attempts
(including ours) to generalize it meaningfully to attributed graphs
and to situations where vertex neighborhoods might be locally similar
rather than identical.

\paragraph{Message Passing:} Inference in graphical models relies on message passing \citep{KolFri09}. This is only exact when the messages are exchanged between maximal cliques in a junction tree, but it is nonetheless often used for approximate inference in general graphs. In these general cases, the algorithm is commonly referred to as {\em loopy belief propagation}. As all algorithms already mentioned in the previous paragraphs, it also consists of iteratively updating some features of each vertex, $\phi(i)$, by incorporating information coming from the neighbors of vertex $i$, commonly referred to as {\em messages}. The outgoing messages of a vertex are obtained by using the local clique potentials in combination with all of its incoming messages, with the exception of the one for which the outgoing message is to be computed. This algorithm can be used to provide features for vertices and there has already been some initial work in that direction (e.g., by \citet{LiLuoJur15}).

All of the algorithms presented in this section can be viewed as a special case of the following iteration:
\begin{align}
\label{eq:background_main_iteration}
	\phi(i) \leftarrow f(\phi(i), 
         & \left\{\phi(j) \mid j \in \mathcal{N}_{\textrm{in}}(i)\right\}, \\
\nonumber 
	&  \left\{\phi(j) \mid j \in \mathcal{N}_{\textrm{out}}(i)\right\})
\end{align}
for some function $f$. We use vertex features of neighbors to compute new vertex features, based on a smartly chosen update function $f$. The locality of this update makes it highly attractive for distributed computation. For PageRank and HITS that function form is explicitly provided by equations \ref{eq:pagerank_iteration} and \ref{eq:hits_iteration}, respectively, and for the Weisfeiler-Lehman iteration we can simply replace $f$ by $h$. Furthermore, we can see how message passing can also fit in this framework by noting that all outgoing messages for a vertex can be computed given all its incoming messages, and these messages could also be passed on appropriately by, for example, tagging each outgoing message with the destination vertex identifier.

\section{PROPOSED APPROACH}

\begin{figure}[t]
    \includegraphics[width=\columnwidth]{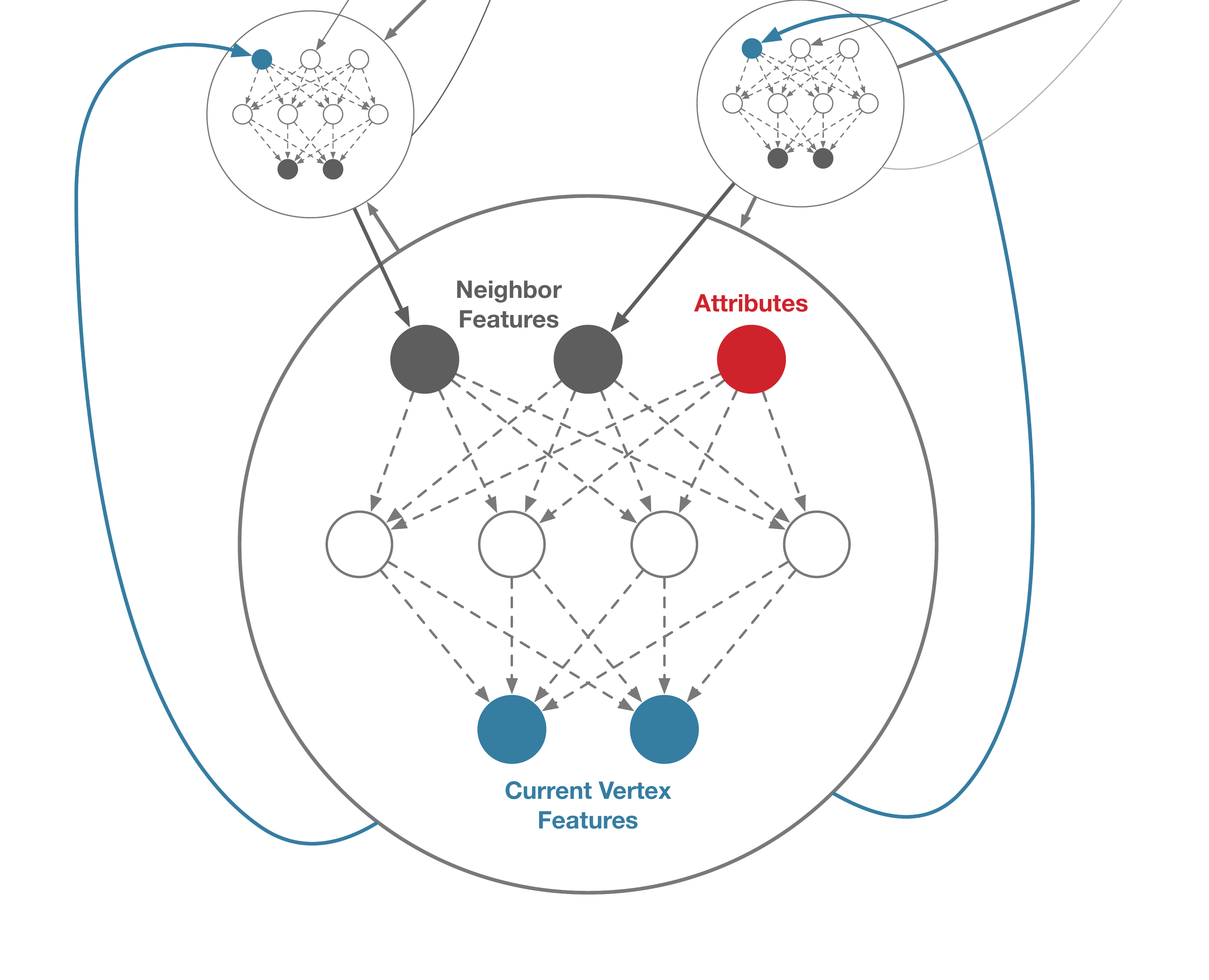}
    \vspace{-2.5em}
    \caption{The Deep Graph algorithm when applied to a small part of an example graph. Big circles represent graph vertices and small circles represent neurons in a deep neural network (DNN). Blue colored neurons correspond to the feature vector of the current vertex and gray colored neurons correspond to the feature vectors of neighbors of this vertex. The red colored neuron represents any attributes that this vertex might have.}
    \label{fig:detailed_graph_figure}
    \vspace{-1em}
\end{figure}

We are now in a position to introduce the key contribution in this work -- the Deep Graph (DG). It consists of the insight that rather than specifying the update equation (shown in equation \ref{eq:background_main_iteration}) {\em manually}, we are at liberty to {\em learn} it based on the tasks at hand. In other words, the update equation now becomes part of the learning system in its own right, rather than being used as a preprocessor, based on heuristics and intuition. In its most general form, it looks as follows:
\begin{equation}
\label{eq:grnn_update_equation}
\begin{split}
	\phi(i) &\leftarrow f(\phi(i), \psi(i) \\ 
	&\qquad\; \left\{\phi(j), \psi(j,i) \mid j \in \mathcal{N}_{\textrm{in}}(i)\right\}, \\
	&\qquad\; \left\{\phi(j), \psi(i,j) \mid j \in \mathcal{N}_{\textrm{out}}(i)\right\};\theta).
\end{split}
\end{equation}
That is, we use inherent vertex and edge attributes in the iteration. Moreover, the system is parametrized by $\theta$ in such a manner as to make the iterations learnable. Note that this approach could also handle learning features over edges in the graph by trivially extending this update equation. However, we are not going to cover such extensions in this work. Now that we have defined the form of the update equation, we need to specify a number of things:
\begin{itemize}[noitemsep,topsep=0pt]
	\item The family of tasks amenable to this iteration.  
	\item The parametric form of the update equation.
	\item Efficient methods for learning the parameters $\theta$.
\end{itemize}
Note that because of the recursive nature of equation \ref{eq:grnn_update_equation}, and given a parametric form for the update function $f$, the computation of the features for all graph vertices consists of an operation that resembles the forward pass of a recurrent neural network (RNN). Furthermore, as we will see in section \ref{sec:parameter_learning}, learning the parameters $\theta$ resembles the training phase of an RNN. Thus comes the name of our approach. An illustration of Deep Graphs is shown in figure \ref{fig:detailed_graph_figure}.

In what follows we omit the vertex and edge attributes $\psi$, without loss of generality, in order to simplify notation.

\subsection{TASKS AMENABLE TO OUR ITERATION}
\label{sec:tasks_amenable}

\begin{figure*}[t]
    \includegraphics[width=\textwidth]{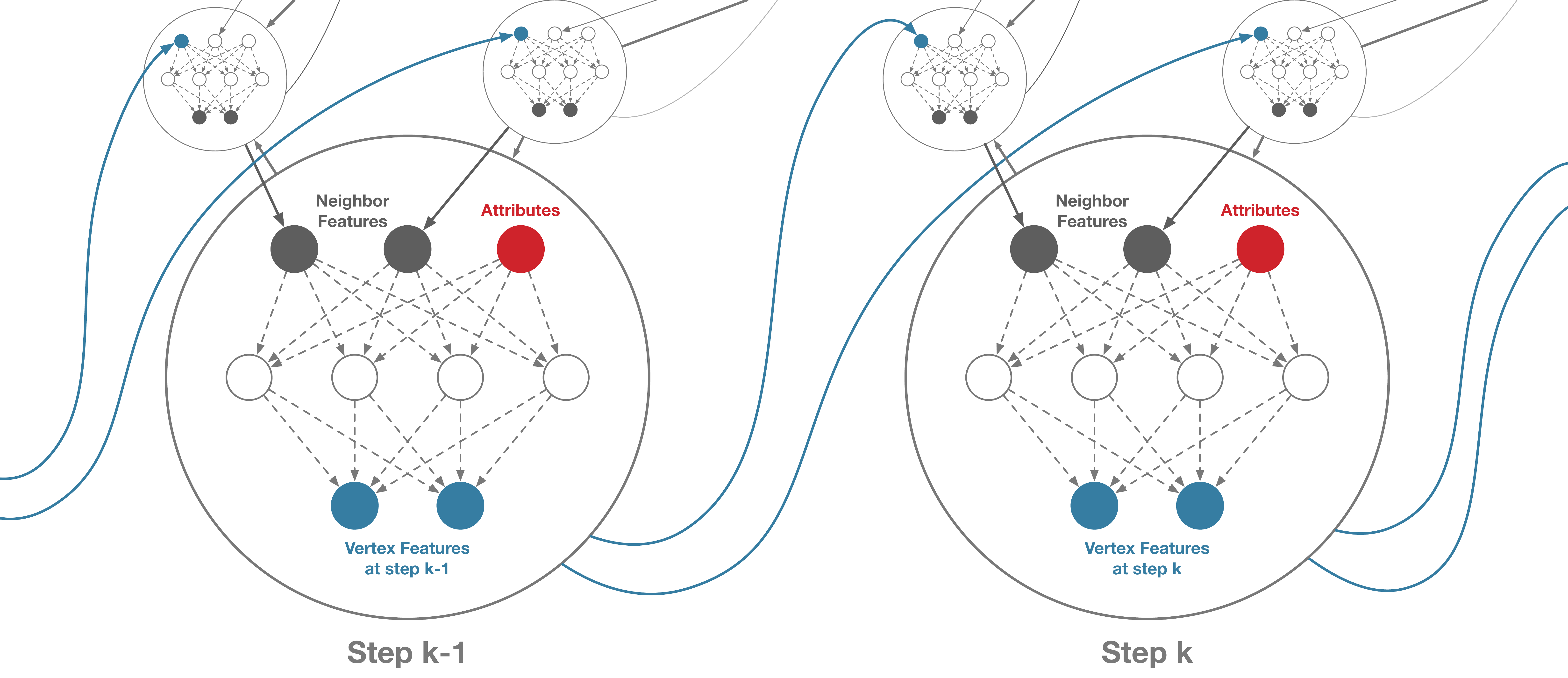}
    \caption{Illustration of the ``unrolling'' of the DGs RNN over the graph structure. This procedure is used for computing the derivatives of the loss function with respect to the model parameters. The ``unrolling'' of the small part of our network presented in figure \ref{fig:detailed_graph_figure}, is shown here.}
      \label{fig:unrolling_graph_figure}
\end{figure*}

Machine Learning usually relies on a set of features for efficient estimation of certain scores. In particular, in the case of estimation on graphs, one strategy is to use vertex features $\phi(i)$ to compute scores $g_i = g(\phi(i), w))$. This is what was used, for example, by \cite{SheBor10} in the context of the Weisfeiler-Lehman algorithm. The authors compute fingerprints of increasing neighborhoods of the vertices $i$ and then use these fingerprints to attach scores to them. For the sake of concreteness, let us consider the following problem on a graph: we denote by $X \triangleq \left\{x_1,\hdots,x_M\right\} \subseteq V$ the training set of vertices for which some labels $Y = \left\{y_1,\hdots,y_M\right\}$ are available (these could be certain vertex attributes, for example). Our goal is to learn a function $g$ that can predict those labels, given the vertex features that $f$ produces and some parameters $w$. We thus want to find a function $g$ such that the expected risk:
\begin{equation}
	R\left\{V \backslash X, g\right\} \triangleq \frac{1}{|V \backslash X|} \sum_{v \in V \backslash X} \mathbb{E}\left\{l(g(v;w), y) \mid v\right\},
\end{equation}
for some loss function $l$, is minimized. In other words, we want to minimize the loss $l$ on vertices not occurring in the training set. Note that there exist quite a few generalizations of this -- for example, to cases where we have different graphs available for training and for testing, respectively, and we simply want to find a good function $g$. Obviously $R\left\{V \backslash X, g\right\}$ is not directly available, but a training set is. Hence, we minimize the empirical estimate $\hat{R}\left\{X, g\right\}$, often tempered by a regularizer $\Omega$ controlling the complexity of $g$. That is, we attempt to minimize:
\begin{equation}
\label{eq:risk_empirical_estimate}
	\hat{R}\left\{X, g\right\} \triangleq \frac{1}{|X|} \sum_{v \in X} l(g(v;w), y) + \Omega\{g\}.
\end{equation}
In the present context, the function $g$ is provided and it is a function of the vertex features $\phi(v)$ parameterized by $w$. The latter denotes our design choices in obtaining a nonlinear function of the features. In the simplest case this amounts to $g(v; w) = \phi(v)^\top w$. More generally, $g$ could be a deep neural network (DNN). In that case, we will typically want to add some degree of regularization in the form of $\Omega\{g\}$ (e.g., via dropout \citep{SriHinKriSutSal14} or via weight decay \citep{Tikhonov43}).

It is clear from the above that a parametric form needs to be provided for $g$ and $l$. Without loss of generality we provide here some example such function forms, for two different kinds of frequently occurring tasks in machine learning.

\paragraph{Regression.} One common task is regression, where the output of $g$ is a real number. That is the case, for example, if we wanted to learn PageRank or HITS (more on this in section \ref{sec:experiments}). In this case $g$ could be modeled a multi-layer perceptron (MLP) \citep{RumHinWil86b} without any activation function in the last (i.e., output) layer\footnote{That is so that its output is unbounded. Note that certain kinds of activation functions could still be used, but that is simply a design choice that does not affect the generality of our method.}, whose input is the feature vector of the corresponding vertex and whose output is the quantity we are predicting (e.g., a scalar for PageRank, and a vector of size two for HITS). Furthermore, $l$ can be defined as the squared $L_2$ norm between the correct output and the MLP output produced by $g$.

\paragraph{Classification.} Another frequently occurring task is vertex classification, where the output of $g$ is a vector of probabilities for belonging to each of a set of several possible classes. In this case we could also use a MLP with a softmax activation function in the output layer. Furthermore, $l$ in this case can be defined as the cross-entropy between the correct class probabilities and the MLP output probabilities produced by $g$.

\subsection{UPDATE EQUATION FORM}
\label{sec:update_equation_form}

Without loss of generality we are going to consider the case where there is no distinction within the set of incoming (or outgoing) edges. We thus need to define a function that is oblivious to order. One may show that when enforcing permutation invariance \citep{GreBorRasSchetal12}, it is sufficient to only consider functions of sums of feature maps of the constituents. For conciseness we drop edge features in the following (the extension to them is trivial). In our case this means that, without loss of generality we can define $f$ as:
\begin{equation}
	f(\phi(i), \phi_{\textrm{in}}(i), \phi_{\textrm{out}}(i)),
\end{equation}
where:
\begin{equation}
	\phi_{\textrm{in}}(i) \triangleq \sum_{j \in \mathcal{N}_{\textrm{in}}(i)} \phi(j),\textrm{ and }
	\phi_{\textrm{out}}(i) \triangleq \sum_{j \in \mathcal{N}_{\textrm{out}}(i)} \phi(j).
\end{equation}
For instance, we could define $f$ as follows:
\begin{equation}
\label{eq:sigmoid_feature_update}
\begin{split}
	&f(\phi(i), \phi_{\textrm{in}}(i), \phi_{\textrm{out}}(i)) \triangleq \\
	&\qquad\sigma\left(W\phi(i)+W_{\textrm{in}}\phi_{\textrm{in}}(i)+W_{\textrm{out}}\phi_{\textrm{out}}(i)+b\right),
\end{split}
\end{equation}
where $\sigma$ is the sigmoid function and $W$, $W_{\textrm{in}}$, $W_{\textrm{out}}$, and $b$ constitute the parameters $\theta$ of $f$ that need to be learned. Even more generally, $f$ could be defined as a MLP or even as a gated unit, such as a Long-Short Term Memory (LSTM) \citep{HocSch97} or a Gated Recurrent Unit (GRU) \citep{Choetal14}, in order to deal with the vanishing gradients problem of taking many steps on the graph \citep{PascanuMB13}.

\subsection{PARAMETER LEARNING}
\label{sec:parameter_learning}

The parameters of our model consist of $\theta$, the parameters of $f$, and $w$, the parameters of $g$. We can learn those parameters by minimizing the empirical estimate of the risk defined in equation \ref{eq:risk_empirical_estimate}, as was already mentioned in section \ref{sec:tasks_amenable}. For that, we are going to need to compute the derivatives of the empirical risk estimate with respect to both $w$ and $\theta$. In the next sections we describe how those derivatives are defined and we then discuss about the optimization method that we use to actually minimize the empirical risk estimate of our model.

\subsubsection{Derivative with respect to $w$}

For the derivative with respect to $w$ we simply have the following:
\begin{equation}
	\frac{\partial\hat{R} \left\{X, g\right\}}{\partial w} = \sum_{v \in X} \frac{\partial\hat{R} \left\{X, g\right\}}{\partial g(v;w)} \frac{\partial g(v;w)}{\partial w},
\end{equation}
where we simply applied the chain rule for differentiation. If $g$ is an MLP as proposed in section \ref{sec:tasks_amenable}, the term $\frac{g(v;w)}{\partial w}$ can be computed by backpropagating the empirical risk gradient \citep{RumHinWil86b}.

\subsubsection{Derivative with respect to $\theta$}
\label{sec:theta_derivative}

For the derivative with respect to $\theta$, the derivation is not as simple due to the recursive nature of $f$. A common approach to deal with this problem is to use backpropagation through structure \citep{Goller96} (based on backpropagation through time \citep{Werbos90}). We will approximate $f$ by considering its application $K$ times, instead of applying it until convergence, where the value $K$ can be set based on the task at hand. That is, the algorithm is going to take a maximum number of $K$ steps in the graph, when computing the features of a particular vertex. This is often referred to as ``unrolling'' an RNN and an example illustration for DGs is shown in figure \ref{fig:unrolling_graph_figure}. At the $k$\textsuperscript{th} step, the feature vector of vertex $i$ is defined as follows:
\begin{equation}
\begin{split}
	\phi^k(i) &= f(\phi^{k-1}(i), \psi(i) \\ 
	&\qquad\; \left\{\phi^{k-1}(j), \psi(j,i) \mid j \in \mathcal{N}_{\textrm{in}}(i)\right\}, \\
	&\qquad\; \left\{\phi^{k-1}(j), \psi(i,j) \mid j \in \mathcal{N}_{\textrm{out}}(i)\right\};\theta).
\end{split}
\end{equation}
Furthermore, the empirical risk is evaluated at the $K$\textsuperscript{th} step only (i.e., after having applied our feature update equation $K$ times and having taken $K$ steps in the graph, for each vertex). Under this approximate setting, the gradient with respect to $\theta$ can be defined as follows:
\begin{equation}
\begin{split}
	\frac{\partial\hat{R} \left\{X, g\right\}}{\partial \theta} &= \sum_{v \in X} \frac{\partial\hat{R} \left\{X, g\right\}}{\partial g(v;w)} \frac{\partial g(v;w)}{\partial \theta}, \\
	&= \sum_{v \in X} \frac{\partial\hat{R} \left\{X, g\right\}}{\partial g(v;w)} \frac{\partial g(v;w)}{\partial \phi^K(v)} \frac{\partial \phi^K(v)}{\partial \theta},
\end{split}
\end{equation}
where for any $k\in\{1,\hdots,K\}$ and any $v \in V$, we know, due to the concept of total differentiation:
\begin{equation}
\begin{split}
	&\frac{\partial \phi^k(v)}{\partial \theta}\!=\!\frac{\partial f(\phi^{k-1}(v),\{\phi^{k-1}(j) \mid j \in \mathcal{N}(v)\};\theta)}{\partial \theta}, \\
	& + \sum_{l=1}^{k-1} \bigg[ \frac{\partial \phi^k(v)}{\partial \phi^l(v)} \frac{\partial \phi^l(v)}{\partial \theta} + \sum_{j \in \mathcal{N}(v)} \frac{\partial \phi^k(v)}{\partial \phi^l(j)} \frac{\partial \phi^l(j)}{\partial \theta}\bigg],
\end{split}
\end{equation}
and for $l<k$, we have that:
\begin{equation}
\begin{split}
	\frac{\partial \phi^k(v)}{\partial \phi^l(j)} &= \frac{\partial \phi^k(v)}{\partial \phi^{k-1}(v)} \frac{\partial \phi^{k-1}(v)}{\partial \phi^l(j)}, \\
	&\qquad + \sum_{i\in\mathcal{N}(v)} \frac{\partial \phi^k(v)}{\partial \phi^{k-1}(i)} \frac{\partial \phi^{k-1}(i)}{\partial \phi^l(j)}.
\end{split}
\end{equation}
These quantities can be computed efficiently by backpropagating the $\frac{\partial \hat{R} \left\{X, g\right\}}{\partial \phi^k(i)}$ for all $i \in V$, as $k$ goes from $K$ to $1$. It is easy to see that the local cost of computing the derivatives is $O(\mathcal{N}(v) + 1)$, hence one stage of the computation of all gradients in the graph is $O(|E| + |V|)$. Thus, the overall complexity of computing the derivatives is $O(K|V| + K|E|)$.

\subsubsection{Optimization}
\label{sec:optimization}

Equipped with the definitions of the empirical risk estimate derivatives with respect to the model parameters, we can now describe our actual optimization approach. An important aspect of the derivative definitions of the last two sections is that, for computing the derivative with respect to the feature vector of a single vertex, the algorithm will very likely have to visit most of the vertices in the graph. Whether or not that happens depends on $K$ and on the connectedness of the graph. However, research has shown that for many kinds of graphs, the algorithm will very likely have to visit the whole graph, even for values of $K$ as small as $3$ \citep{Backstrom2012,journals/corr/abs-1111-4503,DBLP:conf/sdm/KangPST11,Palmer:2002:AFS:775047.775059}. This implies that stochastic optimization approaches that are common in the deep learning community, such as stochastic gradient descent (SGD) and AdaGrad \citep{Duchi:2011:ASM:1953048.2021068}, will likely not be useful for our case. Recent work, such as that by \citet{conf/icml/Martens10}, has shown that curvature information can be useful for dealing with non-convex functions such as the objective functions involved when working with RNNs. For that reason, we decided to use the {Broyden-Fletcher-Goldfarb-Shanno} (BFGS) quasi-Newton optimization algorithm, combined with an interpolation-based line search algorithm that returns a step size which satisfies the strong Wolfe conditions. Furthermore, we set the initial step size for the line search at each iteration by assuming that the first order change in the objective function at the current iterate will be the same as the one obtained in the previous iteration \citep{Nocedal2006}.

\section{EXPERIMENTS}
\label{sec:experiments}

In order to demonstrate the effectiveness of our approach we performed two experiments: a vertex ranking experiment and a vertex classification experiment. Both experiments involve performing a task using only the graph structure as observed information. In the next paragraphs, we provide details on the two evaluations, the data sets we used, and the experimental steup.

\paragraph{Vertex Ranking Experiment.} For this experiment, the goal was to make DGs learn to compute PageRank and HITS scores (as described in section \ref{sec:background}). This is a regression problem, and as a result, we set the form of $g$ to an MLP with a single hidden layer. Thus, $g$ looks as follows:
\begin{equation}
\begin{split}
	h(v) &\triangleq \sigma\left(W_{\textrm{hidden}}\phi(v) + b_{\textrm{hidden}}\right), \\
	g(v) &\triangleq W_{\textrm{out}}h(v) + b_{\textrm{out}},
\end{split}
\end{equation}
where $w \triangleq \{W_{\textrm{hidden}}, b_{\textrm{hidden}}, W_{\textrm{out}}, b_{\textrm{out}}\}$. Furthermore, we selected for the number of hidden units to be twice the size of the feature vector. This is a design choice that did not seem to significantly affect the performance of our model. In order to obtain true labels for the vertices, we ran PageRank and HITS over all graphs, each for $1,000$ iterations. For PageRank, we set the damping factor to be equal to $0.85$, as is often done in practice. The loss function $l$ in this case was defined as follows:
\begin{equation}
	l(g(v;w), y) \triangleq \|g(v;w) - y\|_2^2.
\end{equation}
For simplicity in evaluation, we used the following quantities as scores for the graph vertices: for PageRank, we simply used the PageRank score and for HITS, we used the sum of the hub and the authority scores (note that they are both positive quantities). We evaluated the resulting scores in the following way: we sorted the graph vertices by decreasing true score, and we computed the mean absolute error (MAE) of the DG predictions over the top $10$, $100$, $1000$, and over all vertices, in the sorted vertices list. The reason for this is that in most applications of PageRank and HITS, the highest ranked vertices are most relevant.

\begin{table}[t]
    \caption{Statistics related to the sizes of the data sets we used for our experiments.}
    \label{tab:binary_classification_data_sets}
    \begin{center}
    \begin{tabularx}{\linewidth}{|r@{\hspace{0.3em}}||>{\centering}X|>{\centering\let\newline\\\arraybackslash\hspace{0pt}}X|}
        \hline
        Data Set             & \# Vertices & \# Edges    \\ \hhline{===}
        \textsc{Blogosphere} & $1,490$     & $19,022$    \\ \hline
        \textsc{Dream 5-1}   & $1,565$     & $7,992$     \\ \hline
        \textsc{Dream 5-3}   & $1,081$     & $4,110$     \\ \hline
        \textsc{Dream 5-4}   & $1,994$     & $7,870$     \\ \hline
        \textsc{WebSpam}     & $114,529$   & $1,836,441$ \\ \hline
    \end{tabularx}
    \end{center}
\end{table}

\begin{table*}[t]
\setlength\extrarowheight{2pt}
\setlength\tabcolsep{1.5pt}
    \caption{The results for the PageRank experiment are shown in this table. ``Rank'' corresponds to the number of the highest ranked vertices in the graph that are considered in the evaluation in each experiment (as described in the beginning of section \ref{sec:experiments}). The ``Min'' and ``Max'' rows correspond to the minimum and the maximum PageRank score of those vertices that are considered in the evaluation in each experiment. Our methods' results are highlighted in blue. The numbers correspond to the mean absolute error (MAE) of the predictions (a lower score is better).}
    \label{tab:pagerank_results}
    \vspace{-0.65em}
    \begin{center}
    \begin{tabularx}{\linewidth}{|r@{\hspace{0.3em}}||>{\centering}X|>{\centering}X|>{\centering}X||>{\centering}X|>{\centering}X|>{\centering}X||>{\centering}X|>{\centering}X|>{\centering}X||>{\centering}X|>{\centering}X|>{\centering}X||>{\centering}X|>{\centering}X|>{\centering\let\newline\\\arraybackslash\hspace{0pt}}X|}
        \hline
        Data Set  & \multicolumn{3}{c||}{\thead{\textsc{Blogosphere}\\($\times 10^{-4}$)}} & \multicolumn{3}{c||}{\thead{\textsc{Dream 5-1}\\($\times 10^{-3}$)}} & \multicolumn{3}{c||}{\thead{\textsc{Dream 5-3}\\($\times 10^{-4}$)}} & \multicolumn{3}{c||}{\thead{\textsc{Dream 5-4}\\($\times 10^{-4}$)}} & \multicolumn{3}{c|}{\thead{\textsc{WebSpam}\\($\times 10^{-4}$)}} \\ \hline
        Rank & 10 & 100 & 1000 & 10 & 100 & 1000 & 10 & 100 & 1000 & 10 & 100 & 1000 & 10 & 100 & 1000 \\ \hhline{================}
        Min & $56.2$ & $15.1$ & $0.12$ & $10.9$ & $1.10$ & $0.29$ & $106$ & $12.4$ & $3.20$ & $101$ & $9.60$ & $2.14$ & $6.76$ & $1.82$ & $0.42$ \\ \hline
        Max & \multicolumn{3}{c||}{$117$} & \multicolumn{3}{c||}{$48.0$} & \multicolumn{3}{c||}{$585$} & \multicolumn{3}{c||}{$339$} & \multicolumn{3}{c|}{$15.5$} \\ \hline
        \rowcolor{pale_blue} DG-S & $27.5$ & $0.56$ & $0.06$ & $3.13$ & $0.50$ & \bm{$0.07$} & \bm{$17.6$} & \bm{$5.07$} & $0.59$ & \bm{$21.3$} & \bm{$3.86$} & \bm{$0.71$} & $3.29$ & $1.03$ & $0.19$ \\ \hline
        \rowcolor{pale_blue} DG-G & \bm{$13.4$} & \bm{$0.39$} & \bm{$0.05$} & \bm{$2.46$} & \bm{$0.46$} & \bm{$0.07$} & $22.1$ & $5.28$ & \bm{$0.56$} & $27.0$ & $4.42$ & $0.77$ & \bm{$3.23$} & \bm{$1.01$} & \bm{$0.18$} \\ \hline
    \end{tabularx}
    \end{center}
\end{table*}

\begin{table*}[t]
\setlength\extrarowheight{2pt}
\setlength\tabcolsep{1.5pt}
    \caption{The results for the HITS experiment are shown in this table. ``Rank'' corresponds to the number of the highest ranked vertices in the graph that are considered in the evaluation in each experiment (as described in the beginning of section \ref{sec:experiments}). The ``Min'' and ``Max'' rows correspond to the minimum and the maximum HITS score (i.e., the sum of the authority and the hub scores) of those vertices that are considered in the evaluation in each experiment. Our methods' results are highlighted in blue. The numbers correspond to the mean absolute error (MAE) of the predictions (a lower score is better).}
    \label{tab:hits_results}
    \vspace{-0.65em}
    \begin{center}
    \begin{tabularx}{\linewidth}{|r@{\hspace{0.3em}}||>{\centering}X|>{\centering}X|>{\centering}X||>{\centering}X|>{\centering}X|>{\centering}X||>{\centering}X|>{\centering}X|>{\centering}X||>{\centering}X|>{\centering}X|>{\centering}X||>{\centering}X|>{\centering}X|>{\centering\let\newline\\\arraybackslash\hspace{0pt}}X|}
        \hline
        Data Set  & \multicolumn{3}{c||}{\thead{\textsc{Blogosphere}\\($\times 10^{-2}$)}} & \multicolumn{3}{c||}{\thead{\textsc{Dream 5-1}\\($\times 10^{-2}$)}} & \multicolumn{3}{c||}{\thead{\textsc{Dream 5-3}\\($\times 10^{-2}$)}} & \multicolumn{3}{c||}{\thead{\textsc{Dream 5-4}\\($\times 10^{-2}$)}} & \multicolumn{3}{c|}{\thead{\textsc{WebSpam}\\($\times 10^{-2}$)}} \\ \hline
        Rank & 10 & 100 & 1000 & 10 & 100 & 1000 & 10 & 100 & 1000 & 10 & 100 & 1000 & 10 & 100 & 1000 \\ \hhline{================}
        Min & $23.4$ & $10.4$ & $0.22$ & $23.1$ & $9.66$ & $1.00$ & $23.1$ & $15.0$ & $0.00$ & $52.1$ & $16.6$ & $1.86$ & $12.3$ & $5.53$ & $1.97$ \\ \hline
        Max & \multicolumn{3}{c||}{$35.8$} & \multicolumn{3}{c||}{$94.1$} & \multicolumn{3}{c||}{$132$} & \multicolumn{3}{c||}{$114$} & \multicolumn{3}{c|}{$82.1$} \\ \hline
        \rowcolor{pale_blue} DG-S & $15.2$ & $2.92$ & $0.36$ & \bm{$9.09$} & \bm{$2.39$} & \bm{$0.43$} & $13.1$ & $4.75$ & $0.48$ & \bm{$18.1$} & \bm{$3.40$} & $0.67$ & $7.42$ & $2.98$ & $0.74$ \\ \hline
        \rowcolor{pale_blue} DG-G & \bm{$9.07$} & \bm{$2.73$} & \bm{$0.35$} & $13.1$ & $2.88$ & $0.52$ & \bm{$10.8$} & \bm{$3.21$} & \bm{$0.36$} & $20.2$ & $4.56$ & \bm{$0.64$} & \bm{$7.01$} & \bm{$2.83$} & \bm{$0.73$} \\ \hline
    \end{tabularx}
    \end{center}
\end{table*}

\paragraph{Vertex Classification Experiment.} For this experiment the goal was to make DGs learn to perform binary classification of vertices in a graph. We set the form of $g$ as follows:
\begin{equation}
	g(v) \triangleq \sigma\left(W_{\textrm{out}}\phi(v) + b_{\textrm{out}}\right),
\end{equation}
where $w \triangleq \{W_{\textrm{out}}, b_{\textrm{out}}\}$. For the data sets we used, ground truth vertex labels were provided and were used to evaluate our approach. The loss function $l$ in this case was defined as the cross entropy loss function:
\begin{equation}
	l(g(v;w), y) \triangleq - y \log{p} - (1 - y) \log{(1 - p)},
\end{equation}
where $p \triangleq g(v;w)$. We decided to use this loss function as it has proven more effective than the mean squared error (MSE), in practice, for classification problems utilizing deep learning \citep{golik13}. The metric we used is the area under the precision recall curve (AUC).

\subsection{DATA SETS}

We used the following data sets in our experiments, in order to be able to compare our results with those of \citet{Shervashidze12}, which, to the extent of our knowledge, is the current state-of-the-art in vertex classification (some statistics about those data sets are shown in table \ref{tab:binary_classification_data_sets}):
\begin{itemize}[noitemsep,topsep=0pt]
	\item {\sc Blogosphere}: Vertices represent blogs and edges represent the incoming and outgoing links between these blogs around the time of the 2004 presidential election in the United States. The vertex labels correspond to whether a blog is liberal or conservative. The data set was downloaded from \url{http://www-personal.umich.edu/~mejn/netdata/}.
	\item {\sc Dream Challenge}: This is a set of three data sets that come from the Dream 5 network inference challenge. Vertices in the graph correspond to genes, and edges correspond to interactions between genes. The vertex labels correspond to whether or not a gene is a transcription factor in the transcriptional regulatory network.  We obtained these data sets from \citep{Shervashidze12} through personal communication.
	\item {\sc WebSpam}: This is the WebSpam UK 2007 data set, obtained from \url{http://barcelona.research.yahoo.net/webspam/datasets/}. Vertices in the graph correspond to hosts, and edges correspond to links between those hosts. The vertex labels correspond to whether a host is spam or not. Note that not all of the vertices are labeled in this graph and so, in our experiments, we only train and evaluate on subsets of the labeled vertices. However, the whole graph structure is still being used by our network.
\end{itemize}

\subsection{EXPERIMENTAL SETUP}

\begin{table*}[t]
\setlength\extrarowheight{2pt}
\setlength\tabcolsep{1.5pt}
    \caption{The results for the vertex classification experiment are shown in this table. The numbers correspond to the area under the precision-recall curve (AUC) of the predictions (a higher score is better -- $1$ is a perfect score). Our methods' results are highlighted in blue, and the methods we compare against are described in section 3.7 of \citep{Shervashidze12}.}
    \label{tab:binary_classification_results}
    \vspace{-0.65em}
    \begin{center}
    \begin{tabularx}{\linewidth}{|r@{\hspace{0.3em}}||>{\centering}X||>{\centering}X||>{\centering}X||>{\centering}X||>{\centering\let\newline\\\arraybackslash\hspace{0pt}}X|}
        \hline
        Data Set  & \textsc{Blogosphere} & \textsc{Dream 5-1} & \textsc{Dream 5-3} & \textsc{Dream 5-4} & \textsc{WebSpam} \\ \hhline{======}
        Diffusion & $0.96\pm0.00$ & $0.92\pm0.00$ & $0.72\pm0.04$ & $0.74\pm0.07$ & - \\ \hline
        Laplacian & $0.96\pm0.01$ & $0.83\pm0.01$ & $0.60\pm0.01$ & $0.71\pm0.02$ & - \\ \hline
        $A^2$     & $0.96\pm0.00$ & $0.99\pm0.00$ & $0.83\pm0.01$ & $0.71\pm0.01$ & $0.66\pm0.01$ \\ \hline
        EXACT     & $0.52\pm0.10$ & $0.96\pm0.02$ & $0.90\pm0.06$ & $0.80\pm0.05$ & $0.50\pm0.01$ \\ \hline
        EXACT+NH  & $0.95\pm0.02$ & \bm{$1.00\pm0.01$} & $0.92\pm0.05$ & $0.77\pm0.07$ & $0.66\pm0.03$ \\ \hline
        EXACT+LSH & $0.94\pm0.02$ & \bm{$1.00\pm0.01$} & $0.91\pm0.05$ & $0.78\pm0.07$ & $0.67\pm0.01$ \\ \hline
        LSH WL    & $0.95\pm0.02$ & \bm{$1.00\pm0.01$} & $0.92\pm0.05$ & $0.75\pm0.05$ & $0.67\pm0.01$ \\ \hline
        \rowcolor{pale_blue} DG-S & \bm{$0.98\pm0.01$} & \bm{$1.00\pm0.00$} & \bm{$0.98\pm0.01$} & \bm{$0.98\pm0.01$} & \bm{$0.97\pm0.01$} \\ \hline
        \rowcolor{pale_blue} DG-G & \bm{$0.97\pm0.01$} & \bm{$1.00\pm0.00$} & \bm{$1.00\pm0.01$} & \bm{$0.97\pm0.01$} & \bm{$0.98\pm0.01$} \\ \hline
    \end{tabularx}
    \end{center}
\end{table*}

For our experiments, we performed $10$-fold cross-validation to select model parameters, such as the vertex feature vector size and the maximum number of steps $K$ that our algorithm takes in the graph (as described in section \ref{sec:theta_derivative}). For all experiments, we used feature vectors sizes of $\{1,5,10\}$ and $K$ values of $\{1,2,6,10\}$. We used the best performing parameter setting based on cross-validation on $90\%$ of the labeled data and then we evaluated our methods on the remaining $10\%$. The results of this evaluation are provided in the following section. We tried two options, for the features update equation:
\begin{enumerate}[noitemsep,topsep=0pt]
	\item {\sc Sigmoid}: This the update equation form shown in equation \ref{eq:sigmoid_feature_update}. We denote this method by DG-S in the presentation of our results.
	\item {\sc Gated Recurrent Unit (GRU)}: In order to avoid the well-known problem of vanishing gradients in RNNs \citep{PascanuMB13}, we also tried to use the gated recurrent unit (GRU) of \citep{Choetal14} as our update function form. GRU was initially proposed as a recursion over time, but in our case, this becomes a recursion over structure. Furthermore, what was originally an extra input provided at each time point, now becomes the vector formed by stacking together $\phi_{\textrm{in}}(i)$ and $\phi_{\textrm{out}}(i)$. We denote this method by DG-G in the presentation of our results.
\end{enumerate}
We initialized our optimization problem as follows:
\begin{itemize}[noitemsep,topsep=0pt]
	\item All bias vectors (i.e., vectors labeled with $b$ and some subscript in earlier sections) were initialized to zero valued vectors of the appropriate dimensions.
	\item Following the work of \citet{JozefowiczZS15}, all weight matrices (i.e., matrices labeled with $W$ and some subscript in earlier sections) were initialized to random samples from a uniform distribution in the range $\left[-\frac{1}{\sqrt{n_{\textrm{rows}}}}, \frac{1}{\sqrt{n_{\textrm{rows}}}}\right]$, where $n_{\textrm{rows}}$ is the number of rows of the corresponding weight matrix.
\end{itemize}
For the BFGS optimization algorithm described in section \ref{sec:optimization} we used a maximum number of $1,000$ iterations and an objective function value change tolerance and gradient norm tolerance of $1e-6$, as convergence criteria. We are now in a position to discuss the results of our experiments.

\subsection{RESULTS}

\paragraph{Vertex Ranking Experiment.} The results for the PageRank and the HITS experiments are shown in tables \ref{tab:pagerank_results} and \ref{tab:hits_results}, respectively. It is clear from these results that DGs are able to learn both PageRank and HITS with good accuracy. {\em What is most interesting about these results is that DGs are able to learn to compute the scores by only taking at most $10$ steps in the graph. In fact, for most of our results, the algorithm takes just $6$ steps and produces scores close to PageRank and HITS scores that were computed after applying the corresponding iterations $1,000$ times.} Therefore, even though our approach requires a training phase, it can compute PageRank and HITS scores much more efficiently than the actual PageRank or HITS algorithms can. Furthermore, the results are encouraging because they also indicate that DGs are flexible enough to learn arbitrary vertex ranking functions. Moreover, they can combine attributes of vertices in learning these ranking functions, which could prove very useful in practice.

Also, as expected, DG-G seems to almost always outperform DG-S. This is probably due to the fact that it is better at dealing with the vanishing gradients problem that afflicts DG-S. It would be interesting to apply our models to larger graphs and increase the value of $K$ in our experiments, in order to confirm that this actually the case. We leave that for future work.

We also trained DGs using one graph and apply them on another. We noticed that performance did not significantly differ when our algorithm, having been trained on any Dream challenge graph, was applied on any other of these graphs\footnote{For that reason and due to space constraints, we decided to not include these results in the paper.}. This is intuitive since the graphs corresponding to these data sets are very similar. DGs were also able to do well when using combinations of the Dream challenge data sets and the Blogosphere data set. However, they seemed to perform poorly when using combinations involving the WebSpam data set. The graph in that case is much larger and has significantly different characteristics than the other graphs that we considered, and thus, this result is not entirely unexpected. However, given the rest of our very encouraging results, we believe that we should be able to extend our models, using appropriate forms for functions $f$ and $g$, to generalize well over different graphs.

\paragraph{Vertex Classification Experiment.} Our results for the vertex classification experiment are shown in table \ref{tab:binary_classification_results}. DGs either outperform all of the competing methods by a significant margin, or match their performance when they perform near-perfectly. The most impressive result is for the {\textsc WebSpam} data set. {\em The best competing method achieves an AUC of $0.67$ and DGs are able to achieve an AUC of $0.98$.} That alone is a very impressive and encouraging result for DGs, because it indicates that they can learn to perform diverse kinds of tasks very well, without requiring any task-specific tuning.

\begin{figure}[h!]
\vspace{-1em}
\begin{tikzpicture}
\node[fill=pale_blue,rounded corners=5pt,text width=\columnwidth-14pt, inner sep=7pt] {
\vspace{-1.3em}
\section*{\sffamily\normalsize RESULTS SUMMARY}
\sffamily\footnotesize\justifying
Deep Graphs are able to outperform all competing methods for vertex classification tasks by a significant amount (they achieve an AUC of $0.98$ when the best competing method achieves $0.67$). Furthermore, Deep Graphs can learn to compute PageRank and HITS scores by only applying a vertex update equation $6$ times, as opposed to hundreds or thousands of iterations that the original algorithms require. These are encouraging results that motivate future work for the Deep Graphs framework.\par
};
\end{tikzpicture}
\end{figure}

\section{CONCLUSION}

We have proposed Deep Graphs (DPs) -- a generic framework for deep learning on networks and graphs. Many tasks in machine learning involve learning a representation for the vertices and edges of a graph structure. Deep Graphs can learn such representations, jointly with learning how to perform these tasks. It relies on the notion that many graph algorithms, such as PageRank, Weisfeiler-Lehman, or Message Passing can be expressed as iterative vertex updates. However, in contrast to all these methods, DGs are adaptive to the estimation problem and do not rely on the ingenuity of the designer. Furthermore, they are efficient with the cost of training and deployment being $O(|E| + |V|)$, where $E$ and $V$ are the sets of edges and vertices, respectively. In particular, DGs consist of a recurrent neural network (RNN) defined over the graph structure, where the features of each vertex (or edge) are computed as a function of the neighboring vertices' and edges' features. These features can be used to perform multiple tasks simultaneously (i.e., in a multi-task learning setting) and DGs are able to learn the functions that generate these features and the functions that perform these tasks, jointly. Furthermore, learned features can later on be used to perform other tasks, constituting DGs useful in a transfer learning setting.

We performed two types of experiments: one involving ranking vertices and one classifying vertices. DGs were able to learn how to compute PageRank and HITS scores with much fewer iterations than the corresponding algorithms actually require to converge\footnote{Note here that the iterations for these algorithms have the same complexity as the iterations of our DGs.}. Moreover, they were able to outperform the current state-of-the-art in our vertex classification experiments -- sometimes by a very significant margin. These are encouraging results that motivate future work for the DGs framework.

We are excited about numerous future directions for this work. Our first priority is to perform more extensive experiments with bigger data sets. Then, we wish to try and apply this work to knowledge-base graphs, such as that of \citet{Mitchell:2015wo}, and explore interesting applications of DGs in that direction.

\vspace{-0.7em}


\setlength{\bibsep}{0.0pt}
\bibliography{paper}

\begin{thebibliography}{28}
\providecommand{\natexlab}[1]{#1}
\providecommand{\url}[1]{\texttt{#1}}
\expandafter\ifx\csname urlstyle\endcsname\relax
  \providecommand{\doi}[1]{doi: #1}\else
  \providecommand{\doi}{doi: \begingroup \urlstyle{rm}\Url}\fi

\bibitem[Backstrom et~al.(2012)Backstrom, Boldi, Rosa, Ugander, and
  Vigna]{Backstrom2012}
Lars Backstrom, Paolo Boldi, Marco Rosa, Johan Ugander, and Sebastiano Vigna.
\newblock Four degrees of separation.
\newblock In \emph{Proceedings of the 4th Annual ACM Web Science Conference},
  WebSci '12, pages 33--42. ACM, 2012.

\bibitem[Cho et~al.(2014)Cho, Van~Merri{\"e}nboer, Gulcehre, Bahdanau,
  Bougares, Schwenk, and Bengio]{Choetal14}
Kyunghyun Cho, Bart Van~Merri{\"e}nboer, Caglar Gulcehre, Dzmitry Bahdanau,
  Fethi Bougares, Holger Schwenk, and Yoshua Bengio.
\newblock Learning phrase representations using rnn encoder-decoder for
  statistical machine translation.
\newblock \emph{arXiv preprint arXiv:1406.1078}, 2014.

\bibitem[Duchi et~al.(2011)Duchi, Hazan, and
  Singer]{Duchi:2011:ASM:1953048.2021068}
John Duchi, Elad Hazan, and Yoram Singer.
\newblock Adaptive subgradient methods for online learning and stochastic
  optimization.
\newblock \emph{Journal of Machine Learning Research}, 12:\penalty0 2121--2159,
  July 2011.

\bibitem[Golik et~al.(2013)Golik, Doetsch, and Ney]{golik13}
Pavel Golik, Patrick Doetsch, and Hermann Ney.
\newblock Cross-entropy vs. squared error training: a theoretical and
  experimental comparison.
\newblock In \emph{Interspeech}, pages 1756--1760, Lyon, France, August 2013.

\bibitem[Goller and Küchler(1996)]{Goller96}
Christoph Goller and Andreas Küchler.
\newblock Learning task-dependent distributed representations by
  backpropagation through structure.
\newblock In \emph{In Proc. of the ICNN-96}, pages 347--352. IEEE, 1996.

\bibitem[Gretton et~al.(2012)Gretton, Borgwardt, Rasch, Schoelkopf, and
  Smola]{GreBorRasSchetal12}
A.~Gretton, K.~Borgwardt, M.~Rasch, B.~Schoelkopf, and A.~Smola.
\newblock A kernel two-sample test.
\newblock \emph{JMLR}, 13:\penalty0 723--773, 2012.

\bibitem[Hochreiter and Schmidhuber(1997)]{HocSch97}
Sepp Hochreiter and J{\"u}rgen Schmidhuber.
\newblock Long short-term memory.
\newblock \emph{Neural computation}, 9\penalty0 (8):\penalty0 1735--1780, 1997.

\bibitem[Józefowicz et~al.(2015)Józefowicz, Zaremba, and
  Sutskever]{JozefowiczZS15}
Rafal Józefowicz, Wojciech Zaremba, and Ilya Sutskever.
\newblock An empirical exploration of recurrent network architectures.
\newblock In \emph{ICML}, volume~37 of \emph{JMLR Proceedings}, pages
  2342--2350, 2015.

\bibitem[Kang et~al.(2011)Kang, Papadimitriou, Sun, and
  Tong]{DBLP:conf/sdm/KangPST11}
U~Kang, Spiros Papadimitriou, Jimeng Sun, and Hanghang Tong.
\newblock Centralities in large networks: Algorithms and observations.
\newblock In \emph{SDM}, pages 119--130, 2011.

\bibitem[Kleinberg(1999)]{Kleinberg99}
J.~Kleinberg.
\newblock Authoritative sources in a hyperlinked environment.
\newblock \emph{Journal of the ACM}, 46\penalty0 (5):\penalty0 604--632,
  November 1999.

\bibitem[Koller and Friedman(2009)]{KolFri09}
D.~Koller and N.~Friedman.
\newblock \emph{Probabilistic Graphical Models: Principles and Techniques}.
\newblock {MIT} Press, 2009.

\bibitem[Li et~al.(2015)Li, Luong, and Jurafsky]{LiLuoJur15}
Jiwei Li, Minh-Thang Luong, and Dan Jurafsky.
\newblock A hierarchical neural autoencoder for paragraphs and documents.
\newblock \emph{arXiv preprint arXiv:1506.01057}, 2015.

\bibitem[Low et~al.(2010)Low, Gonzalez, Kyrola, Bickson, Guestrin, and
  Hellerstein]{LowGonKyrBicetal10}
Yucheng Low, Joseph Gonzalez, Aapo Kyrola, Danny Bickson, Carlos Guestrin, and
  Joseph~M. Hellerstein.
\newblock {GraphLab}: A new parallel framework for machine learning.
\newblock In \emph{Uncertainty in Artificial Intelligence}, 2010.

\bibitem[Malewicz et~al.(2010)Malewicz, Austern, Bik, Dehnert, Horn, Leiser,
  and Czajkowski]{Malewicz:2010:PSL:1807167.1807184}
Grzegorz Malewicz, Matthew~H. Austern, Aart~J.C Bik, James~C. Dehnert, Ilan
  Horn, Naty Leiser, and Grzegorz Czajkowski.
\newblock Pregel: A system for large-scale graph processing.
\newblock In \emph{Proceedings of the 2010 ACM SIGMOD International Conference
  on Management of Data}, pages 135--146, 2010.

\bibitem[Martens(2010)]{conf/icml/Martens10}
James Martens.
\newblock Deep learning via hessian-free optimization.
\newblock In \emph{Proceedings of the 30th International Conference on Machine
  Learning, {ICML} 2010, Haifa, Israel, 21-24 June 2010}, pages 735--742, 2010.

\bibitem[Mitchell et~al.(2015)Mitchell, Cohen, Hruschka~Jr, Pratim~Talukdar,
  Betteridge, Carlson, Dalvi, Gardner, Kisiel, Krishnamurthy, Lao, Mazaitis,
  Mohamed, Nakashole, Platanios, Ritter, Samadi, Settles, Wang, Wijaya, Gupta,
  Chen, Saparov, Greaves, and Welling]{Mitchell:2015wo}
Tom~M Mitchell, William~W Cohen, Estevam~R Hruschka~Jr, Partha Pratim~Talukdar,
  Justin Betteridge, Andrew Carlson, Bhanava Dalvi, Matt Gardner, Bryan Kisiel,
  Jayant Krishnamurthy, Ni~Lao, Kathryn Mazaitis, Thahir~P Mohamed, Ndapakula
  Nakashole, Emmanouil~Antonios Platanios, Alan Ritter, Mehdi Samadi, Burr
  Settles, Richard~C Wang, Derry Wijaya, Abhinav Gupta, Xinlei Chen, Abulhair
  Saparov, Malcolm Greaves, and Joel Welling.
\newblock Never-ending learning.
\newblock In \emph{Association for the Advancement of Artificial Intelligence},
  pages 1--9, 2015.

\bibitem[Nocedal and Wright(2006)]{Nocedal2006}
{Jorge} Nocedal and {Stephen J.} Wright.
\newblock \emph{Numerical optimization}.
\newblock Springer series in operations research and financial engineering.
  Springer, 2nd edition edition, 2006.
\newblock ISBN 978-0-387-30303-1.

\bibitem[Page et~al.(1998)Page, Brin, Motwani, and Winograd]{PagBriMotWin98}
L.~Page, S.~Brin, R.~Motwani, and T.~Winograd.
\newblock The {PageRank} citation ranking: Bringing order to the {Web}.
\newblock Technical report, Stanford Digital Library Technologies Project,
  Stanford University, Stanford, CA, USA, November 1998.

\bibitem[Palmer et~al.(2002)Palmer, Gibbons, and
  Faloutsos]{Palmer:2002:AFS:775047.775059}
Christopher~R. Palmer, Phillip~B. Gibbons, and Christos Faloutsos.
\newblock Anf: A fast and scalable tool for data mining in massive graphs.
\newblock pages 81--90, New York, NY, USA, 2002. ACM.

\bibitem[Pascanu et~al.(2013)Pascanu, Mikolov, and Bengio]{PascanuMB13}
Razvan Pascanu, Tomas Mikolov, and Yoshua Bengio.
\newblock On the difficulty of training recurrent neural networks.
\newblock In \emph{Proceedings of the 30th International Conference on Machine
  Learning, {ICML} 2013, Atlanta, GA, USA, 16-21 June 2013}, pages 1310--1318,
  2013.

\bibitem[Rumelhart et~al.(1986)Rumelhart, Hinton, and Williams]{RumHinWil86b}
D.~E. Rumelhart, G.~E. Hinton, and R.~J. Williams.
\newblock Learning representations by back-propagating errors.
\newblock \emph{Nature}, 323\penalty0 (9):\penalty0 533--536, October 1986.

\bibitem[Shervashidze(2012)]{Shervashidze12}
Nino Shervashidze.
\newblock \emph{Scalable graph kernels}.
\newblock PhD thesis, Universit{\"a}t T{\"u}bingen, 2012.

\bibitem[Shervashidze and Borgwardt(2010)]{SheBor10}
Nino Shervashidze and Karsten Borgwardt.
\newblock Fast subtree kernels on graphs.
\newblock In \emph{Neural Information Processing Systems}, 2010.

\bibitem[Srivastava et~al.(2014)Srivastava, Hinton, Krizhevsky, Sutskever, and
  Salakhutdinov]{SriHinKriSutSal14}
N.~Srivastava, G.~Hinton, A.~Krizhevsky, I.~Sutskever, and R.~Salakhutdinov.
\newblock Dropout: A simple way to prevent neural networks from overfitting.
\newblock \emph{The Journal of Machine Learning Research}, 15\penalty0
  (1):\penalty0 1929--1958, 2014.

\bibitem[Tikhonov(1943)]{Tikhonov43}
A.~N. Tikhonov.
\newblock On the stability of inverse problems.
\newblock \emph{Dokl. Akad. Nauk SSSR}, 39\penalty0 (5):\penalty0 195--198,
  1943.

\bibitem[Ugander et~al.(2011)Ugander, Karrer, Backstrom, and
  Marlow]{journals/corr/abs-1111-4503}
Johan Ugander, Brian Karrer, Lars Backstrom, and Cameron Marlow.
\newblock The anatomy of the facebook social graph.
\newblock \emph{CoRR}, abs/1111.4503, 2011.

\bibitem[Weisfeiler and Lehman(1968)]{WeiLeh68}
B.~Weisfeiler and A.~Lehman.
\newblock A reduction of a graph to a canonical form and an algebra arising
  during this reduction.
\newblock \emph{Nauchno-Technicheskaya Informatsia}, 2\penalty0 (9):\penalty0
  12--16, 1968.

\bibitem[Werbos(1990)]{Werbos90}
P.~J. Werbos.
\newblock Backpropagation through time: what it does and how to do it.
\newblock \emph{Proceedings of the IEEE}, 78\penalty0 (10):\penalty0
  1550--1560, 1990.

\end{thebibliography}
\bibliographystyle{plainnat}

\end{document}